\documentclass[journal]{IEEEtranTIE}%journal
\usepackage{graphicx}
\usepackage{cite}
\usepackage{picinpar}
\usepackage{amsmath}
\usepackage{amssymb}
\usepackage{url}
\usepackage[latin1]{inputenc}
\usepackage{colortbl}
\usepackage{soul}
\usepackage{multirow}
\usepackage{pifont}
\usepackage{color}
\usepackage{alltt}
\usepackage[hidelinks]{hyperref}
\usepackage{enumerate}
\usepackage{siunitx}
\usepackage{breakurl}
\usepackage{epstopdf}
\usepackage{pbox}
\usepackage{algorithm}
\usepackage{algpseudocode}
\usepackage{algorithmicx}
\usepackage{calligra}
\usepackage{caption}
\usepackage{graphicx}
\usepackage{float} 
\usepackage{subfigure}
\usepackage{booktabs}
\usepackage{makecell}

\usepackage{amsthm}

\newtheorem{definition}{Definition}

\newtheorem{remark}{Remark}

\begin{document}

\title{Optimization-based Safe Trajectory Planning for Autonomous Ground Vehicle in Multi-Floor Scenarios}

\author{
	\vskip 1em
	
    Zishang Xiang,
	Runda Zhang,
	Runqi Chai, \emph{Senior Member,~IEEE},
        Kaiyuan Chen,\emph{Member,~IEEE}
	Senchun Chai, \emph{Senior Member,~IEEE},
	 and Yuanqing Xia, \emph{Fellow,~IEEE},
	 % and Antonios Tsourdos,\emph{Member,~IEEE}, 

\thanks{Z Xiang, R. Zhang, R. Chai, S. Chai and Y. Xia are with the school of Automation, Beijing
	Institute of Technology, Beijing, China,
	e-mail:(3120240880@bit.edu.cn) (runda.zhang@bit.edu.cn), (r.chai@ieee.org), (chaisc97@bit.edu.cn), (xia\_yuanqing@bit.edu.cn).}
\thanks{ K. Chen is with Vanke School of Public Health, Tsinghua Institute for Healthy China, Tsinghua University, Beijing, China, e-mail: (kaiyuanchen@mail.tsinghua.edu.cn).}
% \thanks{A. Tsourdos is with the School
% 	of Aerospace, Transport and Manufacturing, Cranfield University, UK,
% 	e-mail: (a.tsourdos@cranfield.ac.uk).
%  (Corresponding authors: Runqi Chai)\\}
% <-this % stops a space
}

% 本文针对多楼层服务救援场景提出了一种两阶段的自主轨迹规划框架。第一阶段是一个策略选择阶段，我们基于广义voronoi图和多目标算法提出了一种考虑安全因素的最优出口选择策略。第二阶段是轨迹规划阶段，提出了一个混合优化框架去实现高质量、高效率的轨迹规划。通过选取通道广义voronoi节点，热启动优化问题的求解过程。对于障碍物避障约束，一种自动选择策略被提出去选择相关性强的障碍物约束。同时将障碍物约束凸化来加速数值优化的收敛速率。最后，通过仿真实验验证了所提方法框架的有效性和可行性。

% 通过热启动因子热启动优化算法，同时高效收敛

% AGV的轨迹优化是当下的一大研究热点。本文针对于多层楼的场景提出了一种AGV的轨迹规划框架，这种框架具有快速收敛性。所提框架分为两个部分：任务规划层和轨迹规划层。任务规划层是一个策略选择阶段，一种基于广义voronoi图和多目标算法的任务规划策略被提出去选择每一层楼的楼层出口。轨迹规划层利用基于优化的方法去生成高质量的轨迹，在这一层一个热启动的分层规划框架被设计去保证求解快速收敛。在这其中，针对于复杂的障碍物约束，我们设计了一个相关性约束计算方法用于求解轨迹规划中的障碍物约束。最后，通过仿真实验验证了所提方法框架的有效性和可行性。

\maketitle
\begin{abstract}
The development of trajectory planning strategies for autonomous ground vehicles (AGVs) represents a prevailing research interest within the domain of intelligent transportation systems. This paper introduces a trajectory planning framework tailored for multi-floor scenarios. The framework consists of two main modules: the task planning module and the trajectory planning module. The task planning module involves a strategic selection phase, where a task planning strategy based on generalized voronoi diagrams (GVD) and multi-objective algorithms is proposed to select the floor exits for each floor. The trajectory planning module utilizes optimization-based methods to generate high-quality trajectories, and a warm-started hierarchical planning framework is designed to ensure rapid convergence. Additionally, for handling complex obstacle constraints, a correlation constraint calculation method is designed for reducing obstacle constraints in trajectory planning. Finally, the feasibility and effectiveness of the proposed framework are verified through simulations.
\end{abstract}


\section{Introduction}

% 自动地面车辆由于其灵活性和现有规划控制算法的先进性已经广泛用于公共服务、协同探索、物料运输和火灾救援等场景。AGV的不断发展在给人们生活带来便利的同时还为社会的生产生活创造了巨大的经济效益。AGV系统主要包含环境感知、运动规划、运动控制和通讯系统四个部分。运动规划作为整个系统的上层，主要负责决策。
%路径规划作为运动规划的一个重要组成部分，近年来已经得到了不断地发展。路径规划算法主要分为全局路径规划方法和局部路径规划方法。全局路径规划方法负责全局静态环境下的无障碍路径规划。局部路径规划则是针对动态环境实现局部避障。
%最近几年，路径规划作为运动规划的关键组成部分取得了显著进展。路径规划算法主要分为两大类：全局路径规划和局部路径规划。全局路径规划方法专注于在静态环境下规划无障碍路径，而局部路径规划则致力于在动态环境中实现局部避障。
%现有的全局路径规划算法主要有基于图的方法、基于采样的方法、基于群体智能的算法以及基于人工智能的规划方法。基于图的方法例如A星算法在障碍物已知的情况下，可以快速规划出一条最优路径。基于采样的方法例如RRT*和PRM*算法，利用随机采样策略不断地寻找地图中的最优路径。基于群体智能的算法比如粒子群算法，通过每个粒子的不断搜索，更新个体局部最优和全局最优，从而获得最优路径。基于人工智能的方法主要包含基于深度学习的规划算法和基于强化学习的规划算法。基于图搜索和基于群体智能的方法可以在简单环境下实现一个快速的路径寻优，但是随着环境的复杂度的提高，它们的效率会变得非常低下。基于人工智能的方法可以实现非常高效、智能的路径规划，但是现有的基于人工智能的方法往往不能保证任务的确定完成性。基于采样的算法可以实现效率较高的路径寻优，同时还可以保证方法的概率完备性，但同样也是面临着计算效率低的问题。
%局部路径规划方法包括动态窗口法和弹性带法，这两个方法已经较为成熟的应用在了机器人操作系统中（ROS）。动态窗口法计算复杂度低，但其并不适用于阿克曼模型，这导致在类车机器人上的应用收到了极大的限制。弹性带的方法可适用于各种常见的模型，但是其计算复杂度较大，在复杂环境下可能达不到很好的效果。
% 两类算法近年来都得到了一定的发展，在实际机器人应用中规划方法也基本是全局路径规划方法和局部路径规划方法的组合，即混合路径规划方法。针对上述讲到的问题，如何开发一种高效的路径规划器是一个关键性问题。

% AGV系统主要包含环境感知、运动规划、运动控制和通讯系统四个部分。运动规划作为整个系统的上层，主要负责决策。但是，随着AGV在日常生活中应用的不断拓展，各种复杂环境对其任务执行的表现做出了要求。特别是在多楼层的场景中，AGV的任务执行主要面临着两方面问题：1）如何在多个楼层之间实现高效的任务规划，从而保证路径规划任务的确定性。2）AGV的轨迹规划如何在获得高质量解的同时还能快速收敛。本文的主要目的就是解决这两个问题。

\IEEEPARstart{A}{utonomous} ground vehicles (AGVs) have been widely used in scenarios such as public service, collaborative exploration, material transportation, and fire rescue due to their flexibility and the advancement of existing planning and control algorithms\cite{motionplan1,TIV01,tiv02,li2023embodied}. The continuous development of AGVs brings convenience to people's lives and creates great economic benefits for the production and life of the society. The AGV system mainly consists of four components: environment perception, motion planning, motion control, and communication systems. Motion planning, as the upper layer of the entire system, is primarily responsible for decision-making \cite{runqi_deeplearning}. However, as the application of AGVs expands in daily life, various complex environments impose requirements on their task execution performance\cite{lane-change,tro}. Especially for multi-floor scenarios, AGV task execution primarily faces two issues: 1) How to achieve efficient task planning between multiple floors to ensure the determinism of path planning tasks. 2) How the trajectory planning of AGVs can quickly converge while obtaining high-quality solutions. The main purpose of this paper is to address these two issues.

\subsection{Related Work}
\subsubsection{Motion Planning}
% 无人系统的运动规划一直是非常热门的话题。到目前为止，运动规划方法已经发展出了很多类别，比较热门的方法有基于搜索的方法、基于采样的方法和基于优化的方法。基于搜索的方法包括A星算法及其变体算法，该类算法在栅格化的地图中搜索连接起点到终点的最短路径。其在小规模地图中可以实现较高的搜索效率，但是随着环境规模的增大，搜索难度也一定程度上提高。基于采样的算法，例如快速探索随机树算法和概率路线图算法，通过在无障碍空间内采样和搜索从而找到可行路径。值得注意的是，二者均是找到的较短的路径，但均没有考虑运动学和动力学因约束，导致车辆并不是行驶在一条光滑曲线上。为克服这个问题，有一些工作提出了解决方案。在【1】中，作者在前向扩展时通过求解两点边界值问题来实现节点之间的连接，从而保证轨迹上的非完整约束。这类方法可以以较高效率实现大规模地图下的路径搜索，但是其以概率形式随机采样，收敛速度常常难以保证。

% 基于优化的方法通过将整个运动规划问题指定为一个最优控制问题，通过求解最优控制问题从而实现轨迹的规划。其可以找到满足所制定约束的轨迹，但是其收敛速度一直是研究人员们研究的热点。

The motion planning of unmanned systems has always been a very popular topic. So far, various categories of motion planning methods have been developed, with popular approaches including search-based methods, sampling-based methods, and optimization-based methods. Search-based methods include the A* algorithm and its variants \cite{astar1,TIE_a*}, which search for the shortest path connecting the starting point to the endpoint in a gridded map. These algorithms can achieve high search efficiency in small-scale maps, but as the environment size increases, the difficulty of the search also increases to some extent. Sampling-based algorithms, such as rapidly-exploring random trees (RRT) and probabilistic roadmaps (PRM), sample and search within obstacle-free spaces to find feasible paths \cite{runda_rrt,zhang2024wgit}. It's worth noting that both methods find relatively short paths but do not consider constraints from kinematics, resulting in vehicles not traveling on a smooth curve. To overcome this issue, some work has proposed solutions. In \cite{kin-rrt}, the authors connect nodes during forward expansion by solving two-point boundary value problems to ensure nonholonomic constraints on the trajectory. These methods can efficiently perform path searches in large-scale maps, but their convergence speed is often challenging to guarantee due to the probabilistic nature of the random sampling. In optimization-based approaches, the entire motion planning problem is formulated as an optimal control problem (OCP), and trajectory planning is achieved by solving this OCP \cite{libai_mix,rc_two}. This method can find trajectories that satisfy specified constraints, but the convergence speed has always been a focal point of researchers.

\subsubsection{Optimization-based Method}
% 基于优化的方法中，直接法（Direct Methods）和间接法（Indirect Methods）是两种常见的策略，用于解决最优控制问题或轨迹规划问题。间接法通过构建状态和共轭状态的动力学方程，将最优控制问题转化为一个边值问题（Boundary Value Problem）。通过解决这个边值问题，得到控制参数的最优解，从而实现轨迹规划。间接法通常使用变分法和最优控制理论，涉及到状态变量、共轭状态变量以及庞加莱条件等。直接法直接在控制参数的空间中进行优化,将整个时间段划分为若干个小时间段，然后在这些时间段内对控制参数进行优化，从而获得整体最优解。

In optimization-based methods, indirect methods and direct methods are two common strategies used to solve optimal control problems or trajectory planning problems\cite{opti_indirect,li2024advances}. Indirect methods transform the optimal control problem into a two point boundary value problem (TPBVP) by constructing dynamic equations for the states and conjugate states. By solving this boundary value problem, the optimal solution for control parameters is obtained, thereby achieving trajectory planning. Indirect methods typically involve vocational calculus, optimal control theory, and considerations such as state variables, conjugate state variables, and Pentagons conditions. On the other hand, direct method directly optimizes within the parameter control space, dividing the entire time period into several smaller intervals, and then optimizing the control parameters within these intervals to obtain the global optimal solution.

% 但是对于二维地面上的复杂环境来说，约束多而复杂，直接法更易于进行无人车轨迹的求解。针对于直接法，1等人在【2】中，采用两阶段的优化策略，利用第一阶段的粗轨迹帮助第二阶段优化问题的快速收敛。为了保证最优性，1等人在迭代框架中重建走廊，从而构建轻量级的OCP，实现无人车的最优停车规划。 Tianhao等人考虑停车问题过程之中的机会约束，利用参数连续函数逼近概率约束，进而建立考虑机会约束的OCP。张等人【1】采用基于优化的方法去求解多目标的轨迹规划问题。% 在【1】中，采用了凸化的技术去凸化制定的OCP中的非凸约束，从而使得问题的求解易于收敛。 上述问题的共同特点是根据任务建立完成OCP之后，进行离散化为非线性规划问题，然后采用序列二次规划方法进行数值求解。

% 直接法被很多工作者用来处理面向AGV的各种轨迹规划问题，改进的方法也具有很好的特性。

However, for complex environments on a two-dimensional plane with numerous and intricate constraints, a direct approach is more conducive to solving the trajectory of AGVs. The direct method is widely employed by many researchers to address various trajectory planning problems for AGVs, and improved approaches also exhibit favorable characteristics. In the context of the direct approach, Chai et al. in \cite{runqi_multiphase} employ a two-stage optimization strategy, utilizing the guess from the first stage to expedite the rapid convergence of the second-stage optimization problem. To ensure fast convergence, Li et al. reconstruct corridors within the iterative framework, thus constructing a lightweight optimal control problem (OCP) for achieving optimal parking planning for unmanned vehicles \cite{libai_lightocp}. Liu et al. \cite{tianhao_chance} considered chance-constraints in the parking problem process, using parameterized continuous functions to approximate probability constraints, thereby establishing an OCP that takes chance constraints into account. Zhang et al. \cite{gaochang_fuzzy} employ the direct optimization approach to solve multi-objective trajectory planning problems. 
In \cite{convex1}, convexification techniques were employed to convexify the non-convex constraints in the formulated optimal control problem (OCP), making the problem-solving process more amenable to convergence. The common feature of the above-mentioned approaches is that, after formulating the OCP based on the task, they discretize it into a nonlinear programming (NLP) problem. Subsequently, numerical solutions are obtained using sequential quadratic programming (SQP) methods.

\subsubsection{Multi-floor Scenarios}
% 对于本文所研究的多楼层的AGV应用场景，目前现有一些文献开展了相关研究。在【1】中，针对于多楼层的场景，作者通过引入“虫洞”的概念，表示楼层之间的连接通道。然后利用提出的HRG*算法进行路径规划。在【2】中，作者将智能推荐预测值与迪杰斯特拉算法相结合实现了多楼层的路径规划。但是这些方法缺乏对无人系统运动学约束的考虑。并且考虑包含多个出口的场景，目前现有的方法不能快速作出楼层出口的智能最优决策。
For the multi-floor scenarios studied in this paper, there are existing literatures that have conducted relevant research. In  \cite{multi-floor_mit}, the authors focus on multi-floor scenarios and introduce the concept of "wormholes" to represent connecting passages between floors. They then employ the proposed HRG* algorithm for path planning.  In \cite{multi-floor_2}, the authors combined intelligent recommendation predictions with the Dijkstra algorithm to achieve multi-floor path planning. However, these methods lack consideration for the kinematic constraints of unmanned systems. Additionally, in scenarios involving multiple exits, existing methods cannot quickly make intelligent optimal decisions for floor exits.

\subsection{Contribution}
% 我们提出了一个多层楼环境下的安全、智能和精确的轨迹规划方法为AGV。关键的贡献可以被总结如下：
We have proposed a secure, intelligent, and precise trajectory planning method for AGVs in a multi-floor environment. The key contributions can be summarized as follows:

\begin{enumerate}
% 针对多楼层场景下服务救援问题，一种基于GVD和多目标优化的出口选择策略被设计来为全局路径规划方法提供安全和快速的轨迹规划方案。该策略可以智能地为AGV提供楼层与楼层之间的最优连接通道。
  \item A strategy for exit selection based on GVD and multi-objective optimization is designed to address trajectory planning issues in multi-floor scenarios. This strategy aims to provide a safety and rapid task planning solution for global trajectory planning methods. It intelligently offers AGVs the optimal connecting pathways between floors.
% 本文设计了一种混合优化方法去实现复杂环境下的AGV轨迹规划。混合优化方法中包含对于全局优化的热启动策略，使得优化器快速稳定收敛。
  \item This paper proposes a hierarchical optimization approach to achieve AGV trajectory planning. The hierarchical optimization method includes a warm-start strategy for global optimization, enabling the optimizer to rapidly and stably converge.
% 本文我们对障碍物进行了膨胀，从而考虑一个质点的约束。同时，一种相关性障碍物约束计算的方法被设计以减少规划过程中不必要的障碍物约束。这使得我们的优化问题更容易快速收敛。
  \item A correlation obstacle constraint calculation method is designed to reduce unnecessary obstacle constraints in the planning process. This makes it easier for the optimization problem to converge quickly.

\end{enumerate}

\subsection{Organization}

The remaining sections are organised as follows. Section II presents the mathematical formulation of the trajectory planning problem. Section III presents the overall methodological framework. In Section IV, simulations are conducted to prove the validity of the proposed method. Finally, Section V concludes.

\section{PROBLEM FORMULATION}
% 在本节中，我们将AGV的轨迹规划问题制定为一个最优控制问题。
In this section, we formulate the trajectory planning problem for AGV as an OCP.

\begin{figure}[!htb]
	\centering	
	\includegraphics[width=0.35\textwidth]{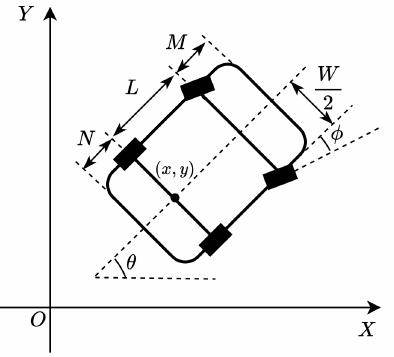}
	\caption{The schematic diagram of AGV kinematic model and the significance of relevant variables.}
	\label{figagv}
\end{figure}
\subsection{AGV Kinematic Model}
% 如下微分公式为本文所考虑的AGV运动学模型（通常被称为自行车模型）：
The following differential formulation shows the kinematic model of the AGV considered in this paper (often referred to as the bicycle model):
\begin{equation}\label{eq1}
    \dot{\boldsymbol{x}}(t) = f(\boldsymbol{x},\boldsymbol{u},t)
\end{equation}
% \begin{equation}\label{eq1}		
% 			\left[ 
% 			\begin{array}{c}
% 				\dot{x}(t) \\
% 				\dot{y}(t) \\
% 				\dot{v}(t) \\
%                 \dot{a}(t) \\
% 				\dot{\theta}(t) \\
% 				\dot{\phi}(t) \\
% 			\end{array} 
% 			\right ] =
% 			\left[ 
% 			\begin{array}{c}
% 				v(t)\cos(\theta(t)) \\
% 				v(t)\sin(\theta(t)) \\
% 				a(t) \\
%                 0 \\
% 				\frac{v(t)\tan(\phi(t))}{l} \\
% 				0 \\
% 			\end{array} 
% 			\right ] +
% 			\left[ 
% 			\begin{array}{c c}
% 				0 & 0 \\
% 				0 & 0  \\
%                 0 & 0 \\
% 				1 & 0  \\
% 				0 & 0  \\
% 				0 & 1  \\
% 			\end{array} 
% 			\right ]
% 			\left[ 
% 			\begin{array}{c}
% 				jerk(t) \\
% 				\omega(t)  \\
% 			\end{array} 
% 			\right ]			
% \end{equation}
% 其中，$ x = \left [ x(t),y(t),v(t),\theta(t),\phi(t) \right ]^{T} $ 表示AGV的状态量。(x,y)表示AGV的位置，v表示速度，\theta表示车辆转向角，\phi表示方向盘转向角。$ u = \left [ a(t),\omega(t) \right ]^{T} $ 表示AGV的控制量，分别代表加速度和方向盘转速。  
where $ \boldsymbol{x} = \left [ x(t),y(t),v(t),a(t),\theta(t),\phi(t) \right ]^{T} $ denotes the states of the AGV. $(x,y)$ denotes the positions, $v$ denotes the velocity, $a$ denotes the acceleration, $\theta$ denotes the vehicle steering angle, and $\phi$ denotes the steering wheel steering angle. $ \boldsymbol{u} = \left [ jerk(t),\omega(t) \right ]^{T} $ denotes the control quantity of the AGV, which represents the jerk and the steering wheel rotation speed respectively. 

\subsection{Boundary Constraints}
% 在多层楼的场景下，AGV在每一层的起点和终点是不同的。由于每一层的规划仍是单独规划的，每次规划时，AGV在起点和终点两个点处需要满足特定的状态，故需要满足如下的边界约束。
In the multi-floor scenario, the starting and ending points of AGVs on each floor are different. Since the planning for each floor remains independent, specific conditions must be met at the starting and ending points of AGVs to satisfy certain states during each planning iteration. Hence, the following boundary constraints need to be fulfilled.
\begin{equation}\label{eq2}
    \boldsymbol{x}(0) = [x_{0},y_{0},0,0,\theta_{0},0]^{T},\boldsymbol{x}(t_{f}) = [x_{f},y_{f},0,0,\theta_{f},0]^{T}
\end{equation}
% 可以看出AGV的起始速度和终止速度均为0，也就是说车辆从零状态开始和停止。
It can be seen that the AGV's start and end speed constraints are both zero, which means that the vehicle starts and stops from a zero state.

\subsection{State and control Constraints}
% AGV的状态量和控制量均受到一定的限制，公式（1）中的变量均在特定的范围内变化，所以状态变量和控制变量受到如下约束：
The state and control values of the AGV are subject to certain constraints, and the variables in Eq. (1) are all varied within a specific range, so the state and control variables are constrained as follows:
\begin{equation}\label{eq3}
    \boldsymbol{x}_{min} \leq \boldsymbol{x}(t) \leq \boldsymbol{x}_{max}
\end{equation}
\begin{equation}\label{eq4}
    \boldsymbol{u}_{min} \leq \boldsymbol{u}(t) \leq \boldsymbol{u}_{max}
\end{equation}
% \begin{equation}\label{eq3}		
% 			\left[ 
% 			\begin{array}{c}
% 				x_{min} \\
% 				y_{min} \\
% 				v_{min} \\
% 				\theta_{min} \\
% 				\phi_{min} \\
% 			\end{array} 
% 			\right ] \leq
% 			\left[ 
% 			\begin{array}{c}
% 				x(t) \\
% 				y(t) \\
% 				v(t) \\
% 				\theta(t) \\
% 				\phi(t) \\
% 			\end{array} 
% 			\right ] \leq
%    			\left[ 
% 			\begin{array}{c}
% 				x_{max} \\
% 				y_{max} \\
% 				v_{max} \\
% 				\theta_{max} \\
% 				\phi_{max} \\
% 			\end{array} 
% 			\right ]
% \end{equation}
% \begin{equation} \label{eq4}	
% 			\left[ 
% 			\begin{array}{c}
% 				a_{min} \\
% 				\omega_{min}  \\
% 			\end{array} 
% 			\right ] \leq
% 			\left[ 
% 			\begin{array}{c}
% 				a(t) \\
% 				\omega(t)  \\
% 			\end{array} 
% 			\right ]	\leq
% 			\left[ 
% 			\begin{array}{c}
% 				a_{max} \\
% 				\omega_{max}  \\
% 			\end{array} 
% 			\right ]    
% \end{equation}
% 其中（·），（·）分别表示变量的最小值和最大值。
where $(\cdot)_{min}$, $(\cdot)_{max}$ denote the minimum and maximum values of the variable, respectively.

\subsection{Obstacle-Avoidance Constraints}
% % 为了保证安全性和任务的可行性，AGV需要躲避环境中的障碍物。因此，障碍物位置与车辆轨迹不能有交集，需要利用如下的限制将车辆与障碍物分隔开：
% To ensure safety and mission feasibility, AGVs need to avoid obstacles in the environment. Therefore, the location of obstacles cannot intersect with the vehicle trajectory and the vehicle needs to be separated from the obstacles using the following restrictions:
% \begin{equation}\label{eq5}	
%     \left\{ {Obstacale} \right\} \cap \left\{ {Trajectory} \right \} = \emptyset
% \end{equation}
% % 上述{obstacle}和{Trajectory}分别表示障碍物和可行轨迹的集合。具体的约束形式随着障碍物的形状的不同而不同。本文所考虑的障碍物为矩形，我们将在下一章针对矩形障碍物设计避障约束。
% The above $\{ Obstacle \}$ and $\{ Trajectory \}$ denote the set of obstacles and feasible trajectories, respectively. The specific form of constraints varies with the shape of the obstacle. The obstacles considered in this paper are rectangular, and we will design obstacle avoidance constraints for rectangular obstacles in the next chapter.

% 为保证AGV不与矩形障碍物发生碰撞，需要制定严格的避障约束。AGV作为一个矩形的车体，可行轨迹需要保证车体不与所有障碍物有交集。但同时这样会带来数量之多的避障约束。为进一步加速OCP收敛速度，我们进一步对障碍物进行了膨胀操作，以减少考虑车体本身碰撞所带来的复杂约束。
To ensure that the AGV does not collide with rectangular obstacles, strict obstacle avoidance constraints need to be formulated.The AGV, as a rectangular vehicle, has a feasible trajectory that ensures that the vehicle does not intersect with all obstacles. But at the same time this will bring a large number of obstacle avoidance constraints. To further accelerate the OCP convergence speed, we further perform an inflation operation on the obstacles to reduce the complexity of the constraints brought about by considering the collision of the vehicle body itself.

\begin{figure}[ht]
	\centering	
	\includegraphics[width=0.48\textwidth]{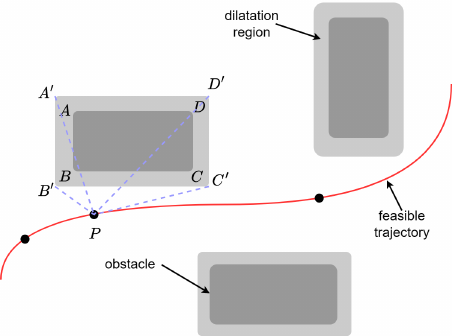}
	\caption{Obstacle inflation and obstacle avoidance constraints.}
	\label{figobs}
\end{figure}

% 膨胀之后可以将车辆视为一个质点。以障碍物A'B'C'D'和点P为例，避免质点进入障碍物区域内，需要制定如下约束：
After the obstacle expansion, the vehicle can be considered as a mass point. Taking obstacles $A'B'C'D'$ and point $P$ as an example, the following constraints need to be formulated to avoid the mass point entering within the obstacle region:
\begin{equation} \label{eq6}
     S_{\Delta PA'B'}+S_{\Delta PB'C'}+S_{\Delta PC'D'}+S_{\Delta PD'A'}>S_{\Box A'B'C'D'}
\end{equation}

% 任意一个凸多边形障碍物均可用如上形式的表达式来进行避障约束的制定。因此对于所有障碍物集合，有如下约束集合：
Any one of the convex polygonal obstacles can be used for obstacle avoidance constraint formulation with an expression of the form as above. Thus for the set of all obstacles, there is the following set of constraints:
\begin{equation}\label{eq20}
    \begin{aligned}
        & S_{\Delta PA_{i}'B_{i}'}+S_{\Delta PB_{i}'C_{i}'}+S_{\Delta PC_{i}'D_{i}'}+S_{\Delta PD_{i}'A_{i}'} \\
        & \ \ \ \ \ \ \ \ \ \ \ \ \ \ \ \  >S_{\Box A_{i}'B_{i}'C_{i}'D_{i}'},i=1,2,...,N_{obs}. 
    \end{aligned}
\end{equation}
% 其中N_{obs}表示考虑的障碍物数量。
where $N_{obs}$ denotes the number of obstacles considered.

\subsection{Objective Function}
% 在本文场景中，为使得整体运行时间最短，我们设计如下的目标函数：
In this paper, to minimise the overall running time, we design the following objective function:
\begin{equation}
    J = t_{f}
\end{equation}

% 因此，整个OCP将由上述目标函数和约束组成，具体OCP形式如下所示：
Therefore, the whole OCP will consist of the above objective function and constraints, and the specific OCP form is shown below:
\begin{equation}\label{eq8}
\begin{array}{cl}
  \text{min} & J= t_{f} \\
  \text{s.t. } & \forall t\in[t_{0},t_{f}]  \\
    & \text{Eq}.(\ref{eq1}) \quad (\text{Dynamic constraints}) \\
    & \text{Eq}.(\ref{eq2}) \quad (\text{Boundary constraints})\\
    & \text{Eq}.(\ref{eq3},\ref{eq4}) \quad (\text{State and control constraints})\\
    & \text{Eq}.(\ref{eq20}) \quad (\text{Obstacle-avoidance constraints})\\
\end{array}
\end{equation}

\section{Overall Trajectory Planning Framework}
% 在本章中，我们将介绍针对多个楼层复杂场景的安全轨迹规划与优化方法。整个规划方法分为两阶段，第一阶段为安全轨迹规划，为整个多层楼环境提供初始解决方案。第二阶段是所设计的混合轨迹优化阶段，通过热启动策略与障碍物处理，确保优化求解快速收敛到局部最优解。
In this section, we present a safe trajectory planning and optimization methodology for multi-floor complex scenarios. The whole planning approach is divided into two phases, the first phase is the safe trajectory planning which provides an initial solution for the whole multi-floor scenarios. The second phase is the designed hierarchical trajectory optimization phase, which ensures that the optimization solution quickly converges to a optimal or near-optimal solution.

\subsection{Multi-floor Safety Planning}
% 正如前面提到的，在火灾救援逃生场景下对AGV进行路径规划，其需要考虑多种因素。本文除了考虑安全性（路径长度）以外还要考虑安全性（路径所在区域的空旷程度）。针对上述特点，本文利用设计的基于广义voronoi图和多目标优化的方法来解决这个问题。
As mentioned earlier, many factors need to be considered when planning the path of AGV in a multi-floor scenario. In addition to optimality (path length), this article also considers safety (the openness of the area where the path is located). Considering the above characteristics, this paper utilises a designed approach based on GVD and multi-objective optimization to solve this problem.
% 在整个环境空间X内，包含一些凸障碍物，这些障碍物的边界存放在集合C中。Xfree中某一点x距离某一障碍物的距离d被定义为

The entire environmental space $\mathbb{X}$ contains a number of convex obstacles whose boundaries are stored in the set $C_{i}$. The distance $d_{i}(x)$ of a point $x$ in $\mathbb{X}_{free}$ from a particular obstacle is defined as
\begin{equation}
    d_{i}(x) = \mathop{min}\limits_{c_{0}\in C_{i}} ||x - c_{0}||
\end{equation}
where $||\cdot||$ is the euclidean distance in $ \mathbb{R} $. The gradient of the $ d_{i}(x) $ is 
\begin{equation}
    \nabla d_{i}(x) = \dfrac{x-c_{0}}{||x - c_{0}||}
\end{equation}
% 其中1表示从c_{0}到x的单位向量。c_{0}表示障碍物C上距离x最近的点。
where $\nabla d_{i}(x)$ represents the unit vector from $c_{0}$ to $x$. $c_{0}$ represents the point on obstacle $C_{i}$ closest to $x$.
% GVD是由一系列二维等距点构成的。我们定义距离障碍物C_{i}和C_{j}的等欧氏距离点如下：

GVD is composed of a series of two-dimensional equidistant points. We define equal Euclidean distance points from obstacles $C_{i}$ and $C_{j}$ as follows:
\begin{equation}
    \mathcal{T}_{ij} = \left\{ x \in \mathbb{X} / (C_{i} \cup  C_{j}): d_{i}(x) = d_{j}(x)     \right \}
\end{equation}

%则二维等距满射曲面，即到达两个物体等距点的集合可定义如下：
Then the two-dimensional equidistant surjective surface, that is, the set of equidistant points reaching two objects, can be defined as follows:
\begin{equation}
    \mathcal{TT}_{ij} = \left \{ x \in T_{ij} : \nabla d_{i}(x) \neq \nabla d_{j}(x)  \right\}
\end{equation}

% 因此，GVD的等距节点可由如下定义：
Therefore, the equidistant face of GVD can be defined as follows:
\begin{equation}
    Fij = \left \{ x \in \mathcal{TT}_{ij} :d_{i}(x) = d_{j}(x) \leq d_{h}(x) ,h \neq i,j    \right \}    
\end{equation}

\begin{figure}[ht]
	\centering	
	\includegraphics[width=0.38\textwidth]{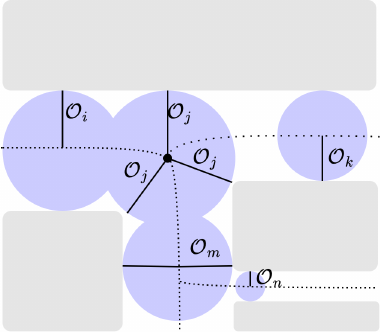}
	\caption{Generalized voronoi diagram processing and openness definition for environments.}
	\label{figgvd}
\end{figure}

% 可以根据上述特点绘制每层楼的广义voronoi图。结果如图2所示。可以直观看到，GVD体现出了地图环境的连通性。根据文献【1】可知，GVD上的路径与最优路径是同伦关系。并且对于某一静态地图，GVD是固定的。综合上述因素，采用GVD路径来反应最优路径的长度。
A generalized voronoi diagram for each floor can be drawn based on the above characteristics. According to the literature \cite{runda_rrt}, it can be seen intuitively that GVD reflects the connectivity of the map. It can be seen that the path on GVD and the optimal path are homotopic. And for a certain static map, GVD is fixed. Based on the above factors, the GVD path is used to reflect the length of the optimal path.

% 对于第二个要考虑的指标环境空旷程度，我们采用如下计算方式来评价某一区域内的拥挤度。
For the second indicator to be considered, space openness, we use the following calculation method to evaluate the openness in a certain area.
\begin{equation}
    \mathcal{O}_{i} = \left \{
    \begin{aligned}
        & 0 , r_{i} \leq \mathcal{E}_{l}   \\
        & r_{i} ,  \mathcal{E}_{l} \leq r_{i} \leq  \mathcal{E}_{u} \\
        & \mathcal{E}_{u}  ,  r_{i} \geq \mathcal{E}_{u} 
    \end{aligned}
    \right.
\end{equation}
% 其中1和2分别表示环境空旷度的最大和最小阈值。3表示GVD节点到障碍物最小距离。
where $\mathcal{O}_{i}$ denotes the openness value at the $i$th generalized voronoi point. ,$ \mathcal{E}_{u} $ and $ \mathcal{E}_{l} $ represent the maximum and minimum thresholds of environmental openness respectively. $ r_{i} $ represents the minimum distance from the GVD node to the obstacle.

% 对于GVD的某一段路径，可以通过如下方式计算该路径的空旷程度：
For a certain path of GVD, the openness of the path $\mathcal{P}_{\mathcal{O}}$ can be calculated as follows:
\begin{equation}
    \mathcal{P}_{\mathcal{O}} = \dfrac{2}{1 + e^{-k(\dfrac{1}{n} \sum_{i = 1}^{n} \mathcal{O}_{i} - x0)}} - 1
\end{equation}
%其中，k表示变化率，x_0表示空旷度值中心值。为表示公平比较，所有空旷度计算值都被上述计算方式映射到了(-1,1)内部。
Here, \( k \) represents the rate of change, and \( x_0 \) denotes the central value of the openness index. For the sake of fair comparison, all calculated values of openness have been mapped to the interval (-1, 1) using the aforementioned method.

%对于找到的所有可行路径集合{\mathcal{P}_{i}},每条路径都有上述两个属性，即路径长度\mathcal{L}_{i}和路径空旷度\mathcal{P}_{\mathcal{O}}.
For all feasible path sets found $\left \{ \mathcal{P}_{i} \right\}$, each path has the above two attributes, namely path length $\mathcal{L}_{i}$ and path openness $\mathcal{P} _{\mathcal{O}}$.

\begin{definition}(Pareto domination)
    %如果对路径集合$ \mathcal{P} $中任意的p1和p2，如果对所有的k = 1，2，...,K,都有f_{k}(p1) \leq f_{k}(p2),则称p2帕累托支配p1.
    If for any $p_{1}$ and $p_{2}$ in the path set $ \mathcal{P} $, for all $k = 1, 2,...,K$, there is $f_{k}(p_{1}) \leq f_{k}( p_{2})$, then $p_{2}$ Pareto dominates $p_{1}$.
\end{definition}

% 根据定义1，在基于广义voronoi图获得的所有可行路径集合后，要分析路径之间的支配关系，从而获得多目标情况下的最优解。
According to Definition 1, after the set of all feasible paths obtained based on the GVD, the dominance relationships between the paths is to be analysed in order to obtain the optimal solution in the multi-objective case.
\begin{figure}[ht]
	\centering	
	\includegraphics[width=0.5\textwidth]{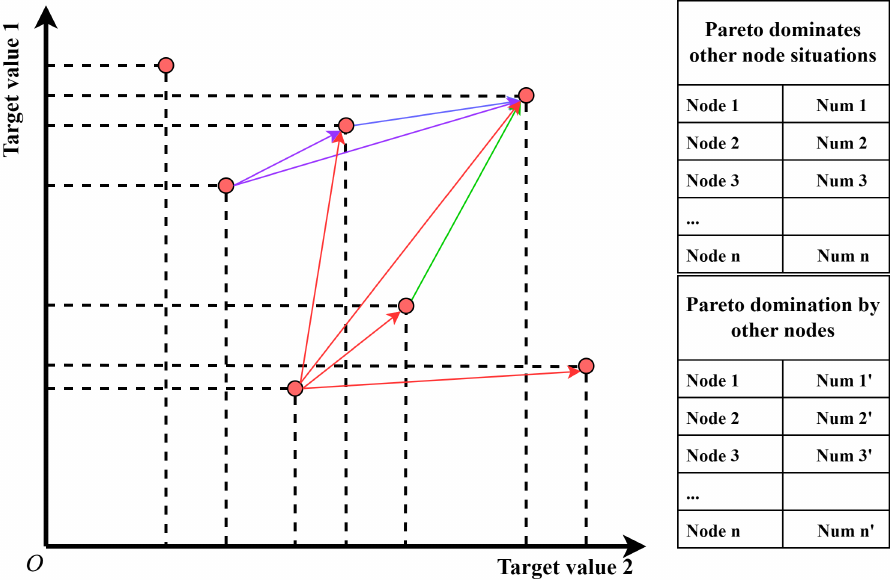}
	\caption{Pareto-based discrete multi-objective selection methods.}
	\label{figpareto}
\end{figure}

% 整个出口选择方案的伪代码如算法1所示。经过算法1计算后，将得到整体任务的执行方案。
The pseudo-code for the entire exit selection scheme is shown in Algorithm 1. After the computation by Algorithm 1, the overall task execution plan will be obtained. The relevant terminologies in the algorithm are explained below:
\begin{itemize}
    \item $ObstacleBoundaryExtraction$ : To extract the boundaries of convex obstacles in the map.
    \item $ GVDProcessing $ :Generate generalized voronoi diagrams for maps.
    \item $ CalculationFeasibleOptions $: Calculate the exit utilization for feasible planning based on each floor exit. (Example: Utilise the first exit on the third floor and the second exit on the first floor to complete the planning task from the third floor to the first floor.)
    \item $ AstarFind $: Solving for feasible paths on GVD maps using the A* algorithm.
    \item $ LengthCalculation $ and $ OpennessCalculation $: Calculate the path length and the openness function of the path, respectively.
    \item $ Pareto $: Calculate the Pareto frontier for a multi-objective task and determine the optimal solution.
\end{itemize}

\begin{algorithm}
	\caption{Exit Selection}\label{al1}
	\begin{algorithmic}[1]
        %\scriptsize
		\Statex \textbf{Input:} Map information set $ \mathcal{M} $ for each floor. % 输入：楼层地图信息集合$ \left \{ \mathcal{M} \right \} $
		\Statex \textbf{Output:} $Floor Exit Sequence$, $GVDmap$.
        \State $ ParameterInitialization() $
        \State $ Boundary = ObstacleBoundaryExtraction(\mathcal{M})  $
        \State $ GVDmap = GVDProcessing(Boundary) $
        \State $ \mathcal{S} = CalculationFeasibleOptions(\mathcal{M}) $
        \State \textbf{for} i = 1: $ \| \mathcal{S} \| $ \textbf{do}
        \State $ \mathcal{P}_{i} = AstarFind(\mathcal{M},\mathcal{S},GVDmap) $
        \State $ \mathcal{L}_{i} = LengthCalculation(\mathcal{P}_{i}) $
        \State $ \mathcal{O}_{i} = OpennessCalculation(\mathcal{P}_{i},\mathcal{M}) $
        \State \textbf{end for}
        \State $ \mathcal{S}_{best} = Pareto(\left\{ \mathcal{L},\mathcal{O}  \right \}) $
        \State \textbf{return} $Floor Exit Sequence$, $GVDmap$.
	\end{algorithmic}
\end{algorithm}

\begin{remark}
    % 本文考虑的多层楼场景下，每一层的规划是独立的。AGV通过与智能电梯通讯从而控制电梯将AGV从当前层载到目标层。即AGV在楼层间的通过过程并不是研究的重点，而整体的全局任务规划和轨迹优化是本项工作的重心。因此，在完成任务规划后，每一层的轨迹规划将是独立的。
    In the multi-floor scenarios considered in this paper, AGVs control elevators to transport them from the current floor to the target floor by communicating with intelligent elevators. Thus, the process of AGVs traversing between floors is not the focus of the study, while the overall global task planning and trajectory optimization are the main emphasis of this work. Therefore, after completing the task planning, the trajectory planning for each layer will be independent.
\end{remark}

\subsection{Hierarchical Trajectory Optimization}
The convergence speed and convergence performance are two crucial metrics considered in solving optimization problems. %It is crucial to be able to ensure that the optimization problem converges to the desired solution. In this paper we design a hierarchical optimization strategy. 
Using the direct method to solve the OCP heavily relies on initial guesses. These initial guesses aid in initiating the OCP-solving process and converging towards nearby solutions. Therefore, generating a high-quality initial guesses is crucial for the convergence of trajectory optimization problems towards the optimum.
% 而生成初始猜测的过程需要注意两点：（1）初始猜测应该全面。初始猜测不应只是路径信息，而应该还包括性能指标时间猜测和轨迹的运动学信息。（2）生成初始猜测的过程应该快速。因此不影响全局优化的速度。本文所提的分层优化框架充分考虑了上述两点。
The process of generating initial guesses should be mindful of two points: (1) Initial guesses should be comprehensive. They should not only consist of path information but also include performance metric time estimations and kinematic information of trajectories. (2) The process of generating initial guesses should be rapid, thereby not impeding the speed of global optimization. The hierarchical optimization framework proposed in this paper takes into full consideration the aforementioned two aspects.
\subsubsection{Hierarchical Optimization Framework}
% OCP的收敛速度和收敛质量均是考虑的指标。能够确保优化问题收敛到理想的解至关重要。本文我们设计了一个混合优化策略。这个优化策略的出发点是考虑到优化问题的时间猜测对于整个OCP求解的重要性。如图所示，较大的时间猜测会导致轨迹收敛到局部最优解，较小的时间猜测可能会使得轨迹难以收敛。因此，所设计的混合优化策略采用两层策略。

% 利用直接法求解轨迹优化问题严重依赖着初始猜测。这个初始猜测帮助最优控制问题在求解过程中从该猜测出发，收敛至附近的解。因此，生成一个高质量的初始解帮助轨迹优化问题收敛至最优解是重要的。通常，高质量的初始猜测应该包括包含运动学信息的状态解猜测以及时间猜测等信息。

%The solution of trajectory optimization problems using direct methods heavily relies on initial guesses. These initial guesses aid in initiating the optimal control problem-solving process and converging towards nearby solutions. Therefore, generating a high-quality initial solution is crucial for the convergence of trajectory optimization problems towards the optimum. Typically, high-quality initial guesses should include state guesses containing kinematic information and time guesses, among other information. %Therefore, the designed hierarchical optimization strategy uses a two-layer strategy.

% The purpose of this optimization strategy is to take into account the importance of the time guess of the optimization problem for the overall OCP solution. As shown in the Fig.5, larger time guesses can cause the trajectory to converge to a locally optimal solution, and smaller time guesses may make it difficult for the trajectory to converge. 

% \begin{figure}[ht]
% 	\centering	
% 	\includegraphics[width=0.5\textwidth]{guesstf.eps}
% 	\caption{Impact of different terminal time guesses on OCP convergence.}
% 	\label{figguess}
% \end{figure}
% 第一层采用低精度优化策略，旨在快速获得全面的初始猜测（包括状态猜测、时间猜测以及运动学信息等）。第一层采用简化运动学模型来保障问题快速收敛。为获得的合适的时间猜测避免轨迹冗余，在低精度优化阶段设置了如下的目标函数：
The first layer employs a low-precision optimization strategy aimed at quickly obtaining comprehensive initial estimations (including state guesses, time guesses, and kinematic information, etc.). Simplified kinematic models are employed in the first layer to ensure rapid convergence of the first layer. To avoid trajectory redundancy and obtain appropriate time guesses, the following objective function is set during the low-precision optimization phase:
% \begin{equation}
%     J = \int_{0}^{t_{f}} | v | dt
% \end{equation}

% 则低精度阶段的最优控制问题形式为如下：

\begin{equation}\label{Sec1eq23}
\begin{array}{cl}
  \text{min} & J = \int_{0}^{t_{f}} | v | dt \\
  \text{s.t. } & \forall t\in[t_{0},t_{f}]  \\
    & \dot{\boldsymbol{x}}_{l}(t) = f(\boldsymbol{x}_{l}(t),\boldsymbol{u}_{l}(t)) \\
    & \boldsymbol{x}_{l}(0) = \boldsymbol{x}_{0},\boldsymbol{x}_{l}(t_{f})=\boldsymbol{x}_{f} \\
    & \boldsymbol{x}_{l} \in [\boldsymbol{x}_{lmin},\boldsymbol{x}_{lmax}],\boldsymbol{u}_{l} \in [\boldsymbol{u}_{lmin},\boldsymbol{u}_{lmax}]\\
    & S_{\Delta PA_{i}'B_{i}'}+S_{\Delta PB_{i}'C_{i}'}+S_{\Delta PC_{i}'D_{i}'}+S_{\Delta PD_{i}'A_{i}'} \\
    & \ \ \ \ \ \ \ \ \ \ \ \ \ \ \ \  >S_{\Box A_{i}'B_{i}'C_{i}'D_{i}'},i=1,2,...,N_{obs}. 
\end{array}
\end{equation}
% 上述目标函数形式是为了使得求解长度较短的轨迹。在较低求解精度下进行离散求解，此时OCP的求解易于收敛。然后将求解结果作为初始猜测热启动求解第二阶段的优化问题。
where $\boldsymbol{x}_{l}$ and $\boldsymbol{u}_{l}$ denote the state and the control of the simplified model. The above objective function form is designed to solve shorter length trajectories. The discrete solution is performed at lower solution accuracy, when the OCP solution is easy to converge. The solution results are then used as an initial guess to warm start solving the second stage of the optimization problem.

% \subsubsection{Warm-start Strategy}
% 在上一小节中，安全规划规划出了每一个楼层考虑安全因素的最短路径，但是并非全局最短路径。因此，在轨迹优化阶段，需要OCP的解收敛到局部最优解。我们设置了一个热启动策略去引导数值优化求解收敛到期望的局部最优解。

%值得注意的是，在求解第一层的优化问题时，一个热启动策略被提出来加速低精度求解阶段的收敛。
% 由于在上一小节中安全规划器规划出了一条由广义voronoi节点连接的路径。因此在低精度求解阶段，广义voronoi节点被均匀采样去引导低精度数值优化收敛。相应的低精度因子如下：
% It is noteworthy that, when solving the optimization problem at the first level, a hot-start strategy is proposed to accelerate the convergence during the low-precision solving phase. In the previous subsection, the safety planner planned the shortest path for each floor considering safety factors, but not the global shortest path. Therefore, in the trajectory optimization phase, it is necessary for the OCP solution to converge to the corresponding local optimal solution. We set up a warm-start strategy to guide the numerical optimization solution to converge to the desired local optimal solution.
It is worth noting that, when solving the optimization problem of the first layer, a warm-start strategy is proposed to accelerate the convergence of the low-precision solving phase. Since in the preceding section, a path connected by generalized voronoi nodes was planned by the safety planner, during the low-precision solving phase, generalized voronoi nodes are uniformly sampled to guide the convergence of low-precision numerical optimization. The corresponding low-precision factors are as follows:

\begin{equation}
    S_{Guess} = [\mathcal{P}^{n}_{best}(1),\mathcal{P}^{n}_{best}(2),...,\mathcal{P}^{n}_{best}(m)]
\end{equation}
% 其中S_{Guess}表示OCP的热启动因子，用于引导OCP收敛到局部最优解。\mathcal{P}^{n}_{best}表示安全规划得到的第n层的路径解。m表示轨迹初始猜测的热启动节点数量。通过在安全规划的路径上均匀cai'dian
where $S_{Guess}$ denotes the warm start factor of the OCP, which is used to guide OCP to converge to the local optimal solution. $\mathcal{P}^{n}_{best}$ denotes the path solution at floor $ n $ obtained by safe planning. $m$ denotes the number of warm start nodes for the initial guess of the trajectory, obtained by uniformly picking $m-2$ points on the safety-planned path.

% 在第二层，将利用第一层求解后的结果作为初始猜测热启动第二层的求解。第二层将利用完成运动学模型并且求解时精度为高精度求解。即第二阶段求解公式（8）.
In the second layer, the results obtained from solving the first layer will be used as initial guesses to hot-start the solving process of the second layer. The second layer will employ a high-precision solving approach, utilizing a completed kinematic model and achieving high-accuracy solutions. This corresponds to solving Eq. (\ref{eq8}) in the second-stage solving process.
For the upper OCP, it can be discretised into a nonlinear planning problem (NLP) and then solved using sequential quadratic programming (SQP).

\subsubsection{Obstacle Constraint Calculation}
% 在全局静态环境已知的条件下，安全规划器已经为全局轨迹做出了指导。部分障碍物约束是不必要的。因此，我们引用了一个障碍物删减策略减少OCP中障碍物的避障约束。
The safety planner has given guidance for the global trajectory under the condition that the global static environment is known. Some of the obstacle constraints are unnecessary. Therefore, we invoke an obstacle deletion strategy to reduce the obstacle-avoidance constraints of obstacles in OCP. 

% \begin{itemize}
    % \item % 首先，利用安全规划得到的圆域进行边界提取得到一组边界点集合{E_{i}}，并提取边界最值x_{min},x_{max},y_{min},y_{max}.
    \textbf{Step 1}: Firstly, a set of boundary point set $\{E_{i}\}$ is obtained by boundary extraction using the circular domain obtained by safety planning, and the boundary maxima $x_{min},x_{max},y_{min},y_{max}$ are extracted.
    \begin{equation}
        \begin{aligned}
         & x_{min} = \min\{ x_{E1},x_{E2},...,x_{Ei},...  \} , \\  
         & x_{max} = \max\{ x_{E1},x_{E2},...,x_{Ei},...  \} , \\
         & y_{min} = \min\{ y_{E1},y_{E2},...,y_{Ei},...  \} , \\
         & y_{max} = \max\{ y_{E1},y_{E2},...,y_{Ei},...  \} 
        \end{aligned}
    \end{equation}
    % \item % 判断每个障碍物与上述边界值是否有交集。需要同时判断横向条件和纵向条件：
        \textbf{Step 2}: Determine whether each obstacle intersects with the above boundary values. Both horizontal and vertical conditions need to be judged:

        horizontal conditions:
        \begin{equation}
            C_{h} = (x_{omin} \leq x_{max}) \wedge (x_{omax} \geq x_{min})
        \end{equation}
        
        vertical conditions:
        \begin{equation}
            C_{v} = (y_{omin} \leq y_{max}) \wedge (y_{omax} \geq y_{min})
        \end{equation}
        % 当横向条件和纵向条件均满足时，将该矩形障碍物列为不必要障碍物，不再考虑其障碍物约束。
        % 其中，xomin、xomax、yomin、yomax分别表示矩形障碍物角点坐标边界值。
        Among these, $x_{omin}$, $x_{omax}$, $y_{omin}$, and $y_{omax}$ respectively represent the boundary values of the coordinates of the corner points of the rectangular obstacle.
        When both the horizontal and vertical conditions are satisfied,  i.e., when $ C_{h} \wedge C_{v} = 1 $, the rectangular obstacle is classified as an unnecessary obstacle and its obstacle constraints are no longer considered.\\
    % \item % 循环检测每个障碍物，最终得到必要的障碍物避障约束。
            \textbf{Step 3}: The cycle detects each obstacle and finally obtains the necessary obstacle-avoidance constraints.

\section{Simulations}

\subsection{Simulation Setup}

% 在本节中，通过模拟多个室内多层场景验证了本文所提方法的有效性和可行性。数值仿真在Windows10的MATLAB R2022b上进行，我们的电脑配置是内存为16G RAM的因特尔i9。所涉及到的场景、算法以及AGV相关的参数总结在了表1中。
In this section, the validity and feasibility of the method proposed in this paper is verified by simulating several indoor multi-layer scenarios. The numerical simulations are performed on MATLAB R2022b in Windows 10, and our computer configuration is an Intel i9 with 16G RAM. the scenarios, algorithms, and AGV-related parameters involved are summarised in Table I.

% 本文所模拟的两个场景如下所示，场景1：每层具有两个出入口的复杂场景，第一层和第三层的两个出口之间被障碍物隔离，第二层的两个出口之间可以被直线连接。用以验证出口选择的有效性。场景2：每层具有四个出入口的复杂场景。用以验证安全规划器对于每个解决方案的目标值评价。所有地图比例均为3：2，障碍物均已进行膨胀操作。
In this paper, two scenarios are simulated as shown below, Scenario 1: a complex scenario with two entrances per level, where the two exits on the first and third levels are separated by a barrier and the two exits on the second level can be connected by a straight line. It is used to verify the validity of the choice of exits. Scenario 2: Complex scenario with four entrances per level. Used to validate the security planner's evaluation of the target value for each solution. All maps are in 3:2 scale and the barriers have been inflated.
\begin{table}[!t]
\caption{Simulation-related Parameters\label{tab:table1}}
\centering
\begin{tabular}{|c||c||c||c|}
\hline
Parameter & Value & Parameter & Value\\
\hline
$x$ & [0,12] & $y$ & [0,8]\\
\hline
$v$ & [-0.6,0.6] & $\theta$ & [-$\pi$,$\pi$]\\
\hline
$\phi$ & [$-33^{\circ}$,$33^{\circ}$] & $a$ & [-0.25,0.25]\\
\hline
$\omega$ & [$-3^{\circ}$,$3^{\circ}$] & $N$ & 0.1\\
\hline
$L$ & 0.4 & $M$ & 0.1\\
\hline
$W$ & 0.3 & $\delta_{low}$ & $10^{-3}$\\
\hline
$\delta_{high}$ & $ 10^{-6} $ & $\mu$ & 0.3\\
\hline
\end{tabular}
\end{table}

\begin{figure}[ht]
	\centering	
	\includegraphics[width=0.48\textwidth]{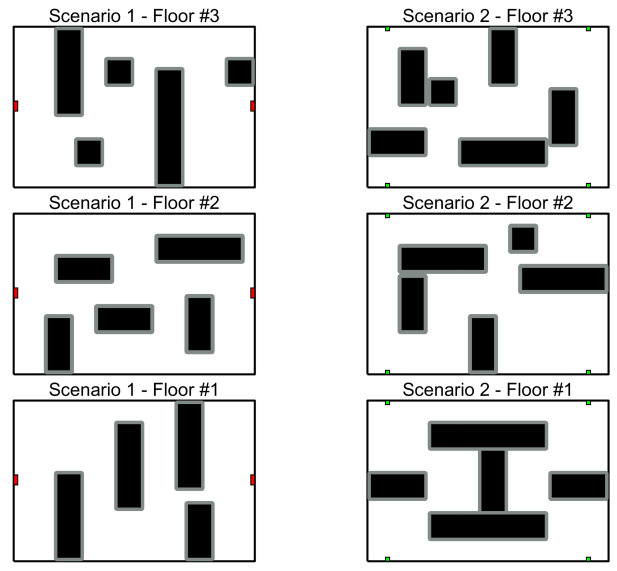}
	\caption{Simulation scenarios.}
	\label{figsc}
\end{figure}

% \begin{figure*}[!ht]
% 	\centering
% 	\subfigure[]{ 
% 		\includegraphics[width=0.4\textwidth]{result_sce1.pdf}}%0.22
% 	\hspace{0in}
% 	\subfigure[]{
% 		\includegraphics[width=0.4\textwidth]{result_sce1.pdf}}
% 	\hspace{0in}
% 	\caption{Simulation scenarios.}
% 	\label{figres1}
% \end{figure*}

\subsection{Results and Discussion}
\begin{figure}[!htb]
	\centering	
	\includegraphics[width=0.45\textwidth]{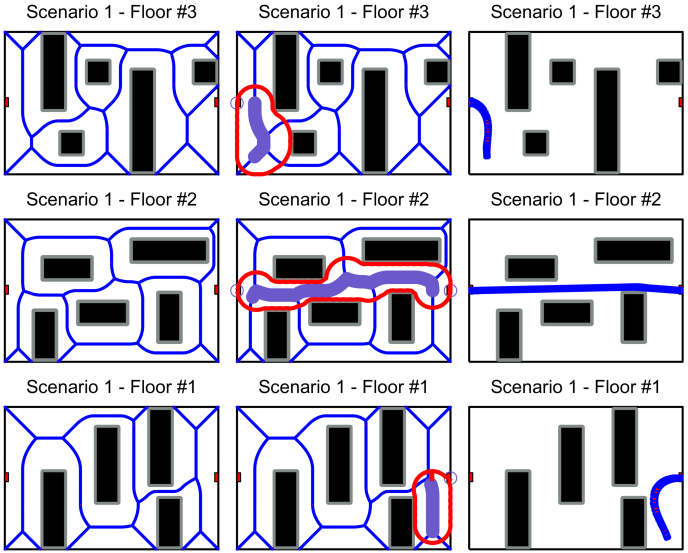}
	\caption{Simulation results in Scenario 1. The exits on each level are designated as exit 1 and exit 2 respectively.}
        % 每一层左右两个出口分别为出口1和出口2.
	\label{figres1}
\end{figure}
% 如图1所示为场景1中利用所提方法框架规划的路径结果。可以看到在已知静态地图的情况下，每层地图的广义voronoi节点被求取（图1最左侧一列）。然后根据任务规划层，得到整体轨迹规划任务的执行方案（图1中间一列紫色部分）。最后利用所提的轨迹优化框架对每一层的轨迹规划任务进行优化求解。图1中最右侧一列为每一层的轨迹结果。可见，在所提的整体框架下，在图1所示场景下完成了整体的规划任务。无人车的轨迹从3层到1层，首先从三层通过左侧出口1到达第二层。然后在二层穿过一层长长的走廊，通过出口2到达第一层最终导航至目标位置。
The path results planned using the proposed method framework in Scenario 1 are shown in Fig. \ref{figres1}. It can be observed that in the case of a known static map, the generalized voronoi nodes for each layer of the map are computed (the leftmost column in Fig. \ref{figres1}). Subsequently, based on the task planning layer, an execution plan for the overall trajectory planning task is derived (the purple section in the middle column of Fig. \ref{figres1}). Finally, the trajectory optimization framework proposed is employed to optimize and solve the trajectory planning tasks for each layer. The rightmost column in Fig. \ref{figres1} displays the trajectory results for each layer. It is evident that within the proposed overall framework, the overall planning task is completed for the depicted scenario in Fig. \ref{figres1}. The trajectory of the unmanned vehicle progresses from level 3 to level 1, initially transitioning from level 3 through exit 1 on the left to reach level 2. Subsequently, on level 2, it traverses a lengthy corridor and passes through exit 2 to reach level 1, ultimately navigating to the target destination.

\begin{figure}[!htb]
	\centering	
	\includegraphics[width=0.45\textwidth]{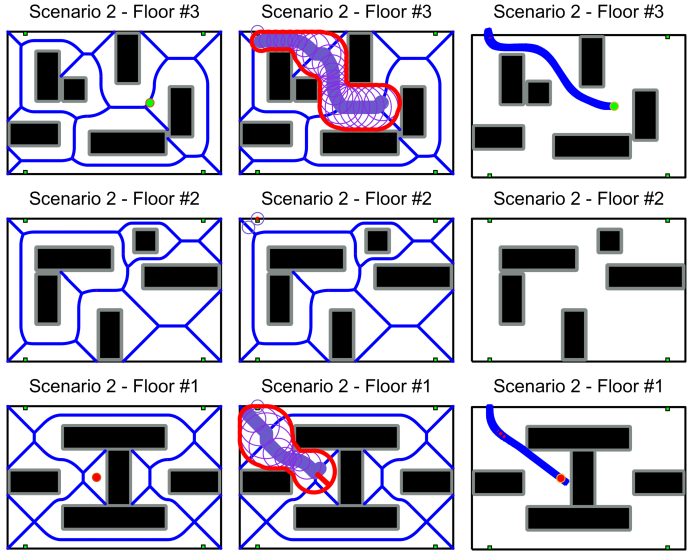}
	\caption{Simulation results in Scenario 2. The upper-left corner is designated as exit 1, the lower-left corner as exit 2, the lower-right corner as exit 3, and the upper-right corner as exit 4.}
        % 每一层的左上角出口为出口1，左下角为出口2，右下角为出口3，右上角为出口4.
	\label{figres2}
\end{figure}

\begin{figure}[ht]
	\centering	
	\includegraphics[width=0.49\textwidth]{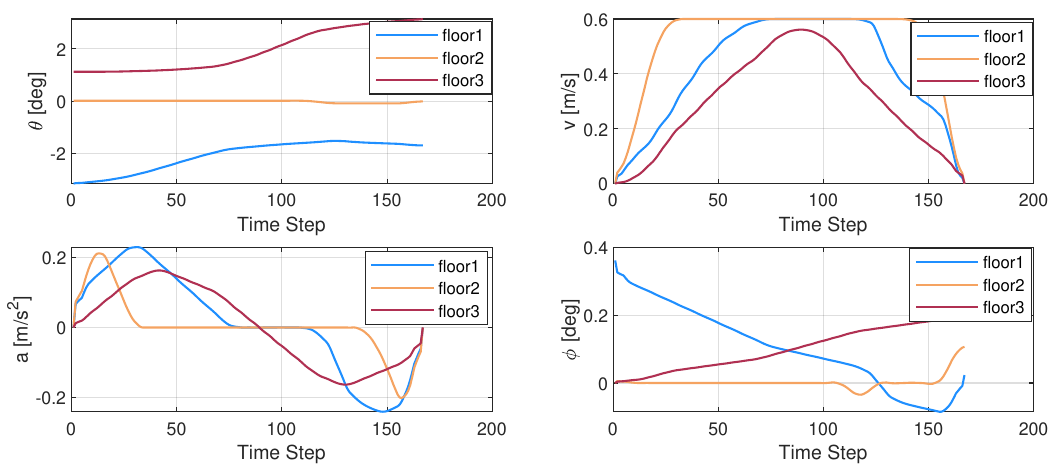}
	\caption{Optimal trajectories for scenario 1.}
	\label{figstateres1}
\end{figure}

\begin{figure}[ht]
	\centering	
	\includegraphics[width=0.49\textwidth]{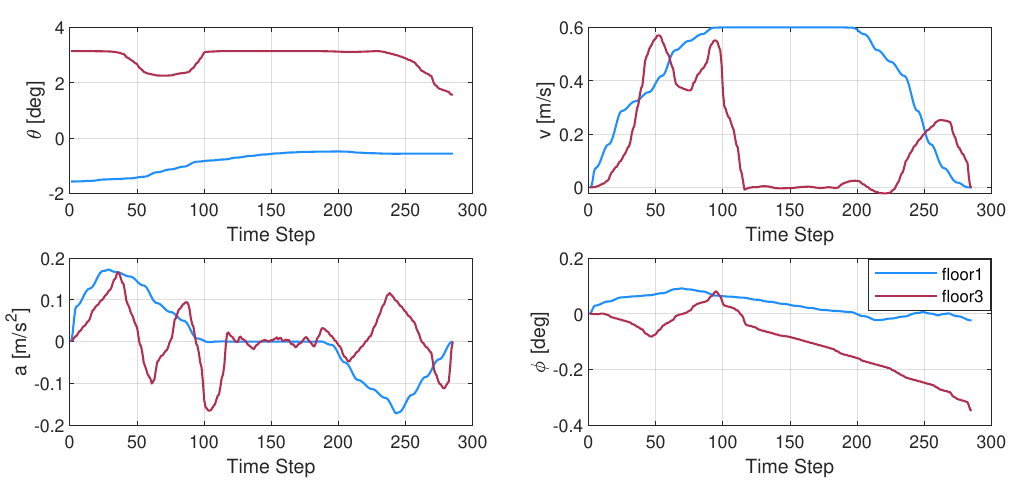}
	\caption{Optimal trajectories for scenario 2.}
	\label{figstateres2}
\end{figure}

% 图2所示为场景2下的轨迹规划结果。场景2中每一层分别包含四个出口，这就涵盖了多样性的任务执行方式。与图1中规划不同的是，场景2中第二层并没有规划轨迹。这是因为从第三层开始车辆通过出口1直接下到了第一层。
The trajectory planning results for Scenario 2 are depicted in Fig. \ref{figres2}. In Scenario 2, each level comprises four exits, thus encompassing diverse task execution approaches. Unlike the planning depicted in Fig. \ref{figres1}, the second level in Scenario 2 does not have trajectory planning. This is because starting from the third level, vehicles descend directly to the first level via Exit 1.
% 两个场景下其余状态量的信息如图9和图10所示。可以看到AGV在行驶过程中速度和加速度等状态量变化连续。结合图7和图8，利用所提的方法框架可以得到平滑的最优解。
The information of the remaining state variables in two scenarios is shown in Fig. \ref{figstateres1} and \ref{figstateres2}. It can be observed that the changes in state variables such as speed and acceleration of the AGV during motion are continuous. Combining Fig. \ref{figres1} and \ref{figres2}, the proposed method framework can yield a smooth optimal solution.

% 表2中所列为

%另外，从表Ⅱ上可以看到两种场景下所有任务规划的可能形式。以场景2中为例，按照本文所提的多目标选择方案，最终选择图2所示的出口方案（1，1）.
Additionally, from Table II, all possible forms of task planning in two scenarios can be observed. Taking Scenario 2 as an example, according to the multi-objective selection scheme proposed in this paper, the final choice (1, 1) is the exit plan as depicted in Fig. \ref{figres2}.

\begin{table}[!t]
\caption{Simulation Case Results of the Proposed Methods\label{tab:tablenew}}
\centering
\begin{tabular}{|c||c||c||c||c|}
\hline
Scenario & Result & Floor1 & Floor2 & Floor3   \\  \hline

\multirow{2}{*}{Scenario1}        & J*    & 10.76s         & 24.92s        & 13.42s  \\  \cline{2-5} 
        & CPU time    &   0.86s        &  2.82s       &  1.03s   \\  % \cline{2-2} \cline{5-7}
\hline
\multirow{2}{*}{Scenario2}        & J*    & 16.94s        & -        & 39.67s  \\  \cline{2-5} 
        & CPU time    &   1.62s        &  -       &  0.92s   \\  % \cline{2-2} \cline{5-7}
\hline
\end{tabular}
\end{table}

\begin{table}[!t]
\caption{Simulation Case Results of the Proposed Methods\label{tab:table2}}
\centering
\begin{tabular}{|c||c||c||c||c||c||c|}
\hline
Scenario & Case & Start & End & length & Openness & Exit  \\  \hline

\multirow{4}{*}{1}        & 1    & \multirow{4}{*}{(1,1)}          & \multirow{4}{*}{(11,1)}         & 38.3962  & 0.6431   & (1,1)  \\  \cline{2-2} \cline{5-7}
        & 2    &           &         & 24.4203  & 0.7467   & (1,2)  \\  \cline{2-2} \cline{5-7}
        & 3    &           &         & 63.7462  & -0.6132   & (2,1)  \\    \cline{2-2} \cline{5-7}
        & 4    &           &         & 27.4537  & 0.7151   & (2,2)  \\ \cline{1-7}
\multirow{16}{*}{2}        & 1    & \multirow{16}{*}{(8,4)}          & \multirow{16}{*}{(5,4)}        & 24.3702  & 0.8356   & (1,1)  \\ \cline{2-2} \cline{5-7}
        & 2    &                  &                & 30.8029  & 0.4688   & (1,2)  \\  \cline{2-2} \cline{5-7}
        & 3    &                  &                & 50.9403  & 0.0120   & (1,3)  \\    \cline{2-2} \cline{5-7}
        & 4    &                  &                & 51.3589  & 0.8590   & (1,4)  \\ \cline{2-2} \cline{5-7}
        & 5    &                  &                & 31.2193  & 0.1406   & (2,1)  \\ \cline{2-2} \cline{5-7}
        & 6    &                  &                & 22.7136  & -0.6237   & (2,2)  \\  \cline{2-2} \cline{5-7}
        & 7    &                  &                & 46.9015  & -0.6604  & (2,3)  \\    \cline{2-2} \cline{5-7}
        & 8    &                  &                & 51.4104  & 0.4534   & (2,4)  \\ \cline{2-2} \cline{5-7}
        & 9    &                  &                & 35.6377  & 0.9596   & (3,1)  \\ \cline{2-2} \cline{5-7}
        & 10    &                  &                & 31.1825  & 0.9041   & (3,2)  \\  \cline{2-2} \cline{5-7}
        & 11    &                  &                & 31.2482  & -0.6501   & (3,3)  \\    \cline{2-2} \cline{5-7}
        & 12    &                  &                & 45.4489  & 0.9540   & (3,4)  \\ \cline{2-2} \cline{5-7}
        & 13    &                  &                & 40.3491  & 0.9847   & (4,1)  \\ \cline{2-2} \cline{5-7}
        & 14    &                  &                & 39.9843  & 0.9628   & (4,2)  \\  \cline{2-2} \cline{5-7}
        & 15    &                  &                & 49.7418  & 0.8289   & (4,3)  \\    \cline{2-2} \cline{5-7}
        & 16    &                  &                & 37.1218  & -0.9388   & (4,4)  \\ \cline{2-2} \cline{5-7}
\hline
\end{tabular}
\end{table}

\subsection{Method Comparison}
% 为了展示所提优化框架的优越性，在本小节我们与其他基于优化的方法进行了对比测试。我们采用本文所提的分层优化框架与如下两种方法进行对比。
To demonstrate the superiority of the proposed optimization framework, in this section, we conducted comparative tests with other optimization-based methods. We compared the hierarchical optimization framework proposed in this paper with the following two methods.

\begin{itemize}
    % 间接法直接进行最优控制问题的求解。
    \item $ M_{1} $: The direct method is employed for the direct resolution of OCPs.
    % 两阶段轨迹优化方法。通过生成路径的初始猜测然后加速最优控制问题的收敛。
    \item $ M_{2} $: A two-stage trajectory optimization method. This method involves generating an initial guess for the trajectory and then accelerating the convergence of the optimal control problem.
\end{itemize}
% 我们以场景2的第三层环境为例对三种方法进行了性能测试。然后来观察在该场景下，三种方法所生成轨迹的优劣。三种方法所生成的轨迹解的最优性能指标tf分别为39.67s，40.92s，41.5s。可以发现利用所提方法生成的轨迹解较另外两种方法更优。三种方法相应解的状态对比图如图8所示。通过图8可以看出，method1相对另外两种方法求解而言状态变化是不同的，其可能是陷入了局部最优解。method0和method1所生成解的状态变化非常相似，但是细致观察可以发现其二者的速度分布不同。二者已经收敛至最优解附近，但是通过性能指标可以看出，所提方法优于method0.

\begin{figure}[ht]
	\centering	
	\includegraphics[width=0.49\textwidth]{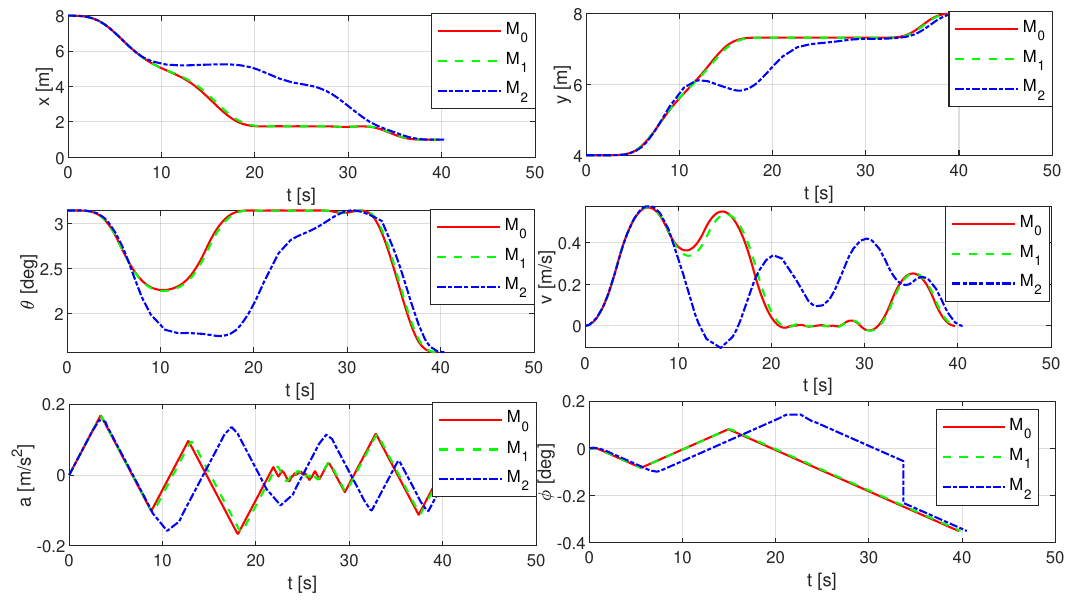}
	\caption{Comparison results of different methods. $M_{0}$ denotes the method proposed in this work.}
	\label{figcompare}
\end{figure}

Performance tests were conducted on three methods using the third-layer environment of scenario 2 as an example. Subsequently, the quality of trajectories generated by these three methods in this scenario was observed. The optimal performance indicators $ t_{f} $ of trajectory solutions generated by the three methods were 39.67s, 40.92s, and 41.5s, respectively. It can be observed that the trajectory solutions generated using the proposed method are superior to the other two methods. The comparative state diagrams of the respective solutions of the three methods are shown in Fig. \ref{figcompare}. From Fig. \ref{figcompare}, it can be seen that the state transitions of method $ M_{1} $ are different from the other two methods, possibly indicating convergence to a local optimum. The state transitions of solutions generated by method $ M_{0} $ and method $ M_{2} $ are very similar, but a detailed observation reveals differences in their velocity distributions. Both methods have converged to near-optimal solutions, but based on performance indicators, it can be concluded that the proposed method outperforms method0.

% 为了验证本文所提优化框架的稳定性，我们进行了蒙特卡洛稳定性测试。在对起点和终点进行随机扰动变化后，进行大量的试验测试，统计方法成功的占比。
To verify the stability of the optimization framework proposed in this paper, we conducted Monte Carlo stability tests. After introducing random perturbations to the starting and ending points, a significant number of experimental trials were conducted to assess the success rate of the statistical methods.

% 扰动变量的分布如表III所示。每次测试施加随机的初始变量扰动，然后对三种方法进行初始化设置，利用三种方法进行求解。可以得到三种方法下的求解成功率分别为82.6%，72.5%，32.3%.从而可以看出所提方法在进行轨迹求解时面对扰动是具有一定的鲁棒性的。
The distribution of the perturbation variables is presented in Table III. Each test applies random initial variable perturbations, followed by initialization settings for three methods, utilizing them for solving. The success rates under the three methods are 89.6\%, 72.5\%, and 32.3\% respectively. Thus, it can be observed that the proposed method exhibits a certain robustness when faced with perturbations during trajectory solving.

\begin{table}[!htb]
\caption{Monte-carlo Parameters\label{tab:table3}}
\centering
\begin{tabular}{|c||c||c||c||c||c|}
\hline
State   & Distribution & 3-$\sigma$ range & State   & Distribution & 3-$\sigma$ range\\
\hline
$x_{0}$ & Gaussian     & 0.25    & $y_{0}$ & Gaussian     & 0.25         \\
\hline
$\theta_{0}$ & Gaussian     & 0.16       &   $\phi_{0}$ & Gaussian     & 0.16    \\
\hline
\end{tabular}
\end{table}

% 综上所示，利用所提方法框架进行轨迹优化问题的求解不仅可以获得较高质量的轨迹解，而且还具有一定的鲁棒性。这展现了所提方法的优越性。
Based on the above, utilizing the proposed method framework for solving trajectory optimization problems not only yields trajectories of higher quality but also demonstrates a certain level of robustness. This showcases the superiority of the proposed method.

\section{CONCLUSION}
% 在本项工作中，多层场景下的轨迹规划与优化问题被研究。首先，基于广义voronoi图和帕累托多目标选择方案的安全规划方法被设计去选择全局的可行方案。在此基础上，一个分层递进的优化框架被提出来解决每一层实际的轨迹优化问题。在这个分层优化框架中，首先我们采用利用简化模型进行低精度问题求解，然后利用低精度解去加速第二层的精确优化问题的收敛。最后，通过大量的数值仿真观察到，所提方法可以实现多层楼之间整体的全局轨迹规划，并且局部轨迹优化框架具有较高的收敛率和鲁棒性。
In this study, trajectory planning and optimization problems in multi-layered scenarios are investigated. Initially, a safety planning approach based on generalized voronoi diagrams and Pareto multi-objective selection schemes is designed to select global feasible solutions. Building upon this, a hierarchical progressive optimization framework is proposed to address the actual trajectory optimization problems at each layer. In this hierarchical optimization framework, we first employ simplified models to solve low-precision problems, and then utilize the low-precision solutions to accelerate the convergence of the precise optimization problems at the second layer. Finally, through extensive numerical simulations, it is observed that the proposed approach can achieve holistic global trajectory planning among multiple floors, and the local trajectory optimization framework exhibits high convergence rates and robustness.

% 在接下来的工作中，我们将考虑将所提算法应用到实际的多层楼的实际场景中。同时，在轨迹规划阶段，我们考虑加入环境不确定性的因素，来更加逼近实际场景中的轨迹规划。
In the future work, we will contemplate the application of the proposed algorithm to real-world scenarios involving multiple floors. Simultaneously, during the trajectory planning phase, we intend to incorporate factors of environmental uncertainty to achieve a closer approximation to trajectory planning in real-world scenarios.

\bibliographystyle{IEEETranTIE}
\bibliography{reference.bib}\ %IEEEabrv instead of IEEEfull

\end{document}